\documentclass[letterpaper, 10 pt, conference]{ieeeconf}  
\pdfoutput=1

\IEEEoverridecommandlockouts                              

\usepackage[utf8]{inputenc}
\usepackage[normalem]{ulem}
\usepackage{graphicx}
\usepackage{color}
\usepackage{multirow}
\graphicspath{ {Figs/} }
\usepackage{float}
\usepackage{amsmath} 
\usepackage{amssymb}  
\usepackage{colortbl}
\usepackage{cite}
\usepackage{comment}
\usepackage{soul}

\DeclareMathOperator*{\argmin}{arg\,min}

\def\traj{\tau}
\def\obs{\mathbf{o}}
\def\act{\mathbf{a}}
\def\conf{\mathbf{c}}
\def\state{\mathbf{s}}
\def\params{\theta}
\def\bev{\mathbf{\phi}}
\title{\LARGE \bf
Driving Through Ghosts: Behavioral Cloning with False Positives
}

\author{Andreas B\"uhler$^{1,2}$, Adrien Gaidon$^{2}$, Andrei Cramariuc$^{1}$, Rares Ambrus$^{2}$, Guy Rosman$^{2}$, and Wolfram Burgard$^{2}$%
\thanks{\footnotesize $^{1}$Autonomous Systems Lab (ASL), ETH Zurich, Switzerland
{\tt\footnotesize andreas@abuehler.li, andrei.cramariuc@mavt.ethz.ch}}%
\thanks{\footnotesize $^{2}$Toyota Research Institute (TRI), USA
{\tt\footnotesize first.last@tri.global}}%
}

\begin{document}
\maketitle
\thispagestyle{empty}
\pagestyle{empty}

\begin{abstract}
Safe autonomous driving requires robust detection of other traffic participants. 
However, robust does not mean perfect, and safe systems typically minimize missed detections at the expense of a higher false positive rate. 
This results in conservative and yet potentially dangerous behavior such as avoiding imaginary obstacles.
In the context of behavioral cloning, perceptual errors at training time can lead to learning difficulties or wrong policies, as expert demonstrations might be inconsistent with the perceived world state.
In this work, we propose a behavioral cloning approach that can safely leverage imperfect perception without being conservative.
Our core contribution is a novel representation of perceptual uncertainty for learning to plan.
We propose a new probabilistic birds-eye-view semantic grid to encode the noisy output of object perception systems.
We then leverage expert demonstrations to learn an imitative driving policy using this probabilistic representation.
Using the CARLA simulator, we show that our approach can safely overcome critical false positives that would otherwise lead to  catastrophic failures or conservative behavior.
\end{abstract}

\section{Introduction}

\begin{figure}[t!]
\centering
\includegraphics[width=0.48\textwidth]{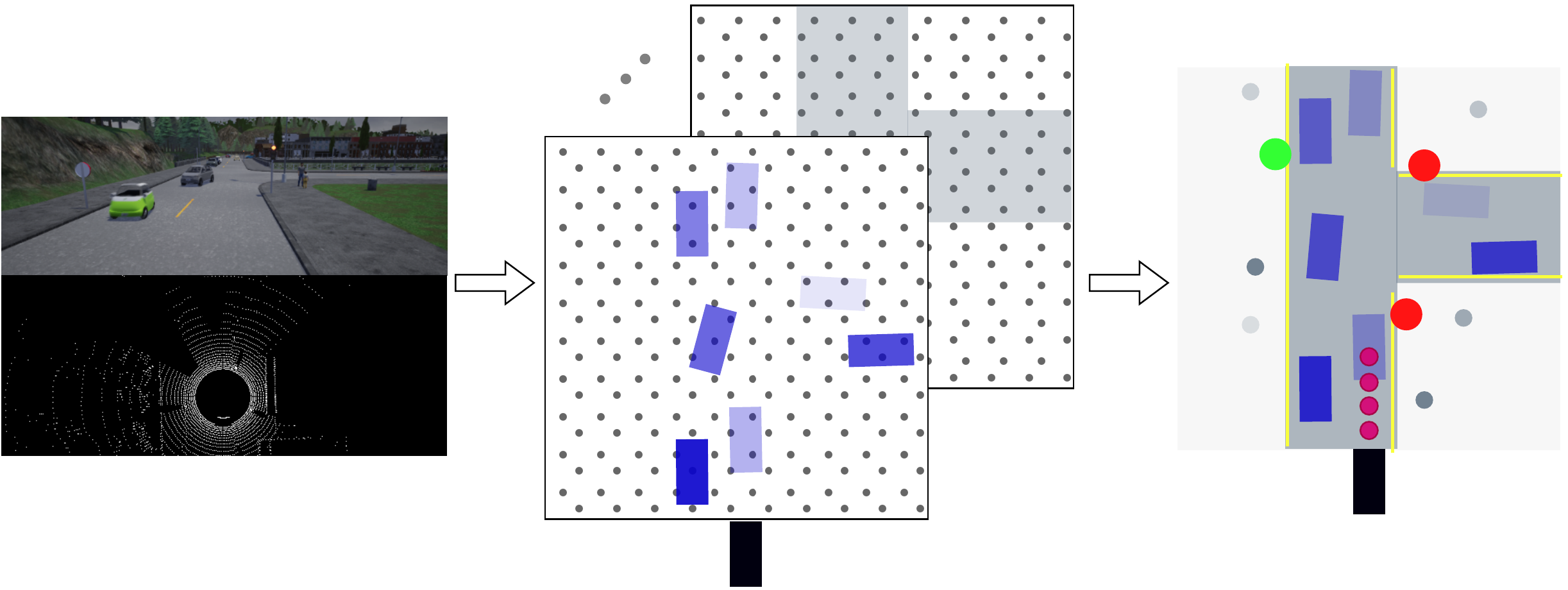}
\caption{We propose a probabilistic birds-eye-view semantic representation, \textit{Soft BEV}, for imitation learning under perceptual uncertainty. It enables learning safer policies that can ignore false positives.}
\label{fig:teaser}
\vspace*{-4mm}
\end{figure}

Safety is the most critical concern when building autonomous robots that operate in human environments.
For autonomous driving in particular, safety is a formidable challenge due to high speeds, rich environments, and complex dynamic interactions with many traffic participants, including vulnerable road users.
Sensorimotor imitation learning methods tackle this problem by learning end-to-end from demonstrations~\cite{Pomerleau1988,LeCun2005driving,Bojarski2016nvidiadriving,Codevilla2018, Wang18}.
Although scalable, these approaches suffer from generalization issues~\cite{cilrs2019iccv}, and thus lack statistical evidence of safe behavior.
In contrast, modular approaches plan using perceptual abstractions~\cite{chen2015deepdriving,muller2018driving,Sauer2018conditional}, which leads to good generalization properties~\cite{Zhou2019DoesCV}.
%
However, modular systems often suffer from error propagation. 
Upstream perceptual errors may yield incorrect abstractions, which then, when consumed by the downstream planner, can lead to critical failures.
%
Although robust sensor fusion can reduce perceptual errors, it cannot totally eliminate them.
Furthermore, in the case of imitation learning, perceptual errors at training time can lead to learning difficulties, as expert demonstrations might be inconsistent with the world state perceived by the vehicle~\cite{Silver2010navigation}.
Consequently, learning and deploying modular imitative policies in the real world requires modeling perceptual uncertainty during both learning and inference.
Planning under uncertainty is a long-standing research topic, but most approaches do not account for errors in \textit{perceptual inputs}, focusing instead on uncertainty in the dynamics~\cite{RossB12,lee2017gp,harrison2020adapt}, future trajectories~\cite{rhinehart2019precog, ivanovic2019trajectron,huang2019uncertainty,huang2020carpal}, model weights~\cite{cui2019udagger}, demonstrations~\cite{wu2019imitation}, or the cost function~\cite{rhinehart2020deep}.
As modern perception systems rely on deep neural networks for which uncertainty modeling is an active area of research~\cite{kendall17uncertainty, Seg2020AGF}, it is still unclear how to properly model and use those uncertainties in downstream components like a planner. 
In this work, we propose a behavioral cloning algorithm that uses a non-parametric representation of an uncertain world state as predicted by typical perception systems.
In particular, we focus on critical perception mistakes in the form of detection false positives, as this is typically the main failure mode of safe systems biased towards negligible false negative rates (as false negatives are irrecoverable in mediated perception approaches).
%
Our data-driven approach bypasses modeling steps typically performed when formulating the problem in a classical manner, i.e. as a POMDP.
The major drawback of such classical methods is the computational complexity, where real-time performance can only be achieved through major simplifications of the problem, often ignoring the typical failure modes of perception in the belief approximation.
%
%
\\
The \textbf{main contribution} of our work is a novel input representation that combines predicted visual abstractions and scalar confidence values by convolving them in a discrete top-down birds-eye-view grid, which we call a \textit{Soft BEV} grid.
%
%
Our \textit{Soft BEV} captures perceptual uncertainty across the full scene, and is able to exhibit probabilistic patterns to deal better with inconsistencies. 
\\
Our \textbf{second contribution} is to show the usage of this representation in a Deep Learning planning model, trained to imitate expert demonstrations, which can be inconsistent with the world state perceived by the vehicle (cf.~Fig.~\ref{fig:teaser}). 
Specifically, our policy network learns when to trust perception or not, thus recovering imitative policies that safely avoid excessively conservative behavior in the presence of perception errors. 
We use the CARLA simulator~\cite{Dosovitskiy2017} to demonstrate the effectiveness of our proposed approach.

%



\section{Related work}

Traditional approaches for planning under uncertainty often require efficient representations of state uncertainty. This includes methods from trajectory optimization~\cite{platt2010belief, patil2015scaling,xu2014motion,todorov2005generalized,van2017motion} and POMDP-based techniques~\cite{bai2015intention, ahmadi2019riskaverse}. Other approaches focus on specific representations of uncertainty. For example, Richter \emph{et al.}~\cite{richter_isrr15} presented a non-parametric Bayesian approach to learn collision probabilities in the context of navigating unknown environments.

In robotics, there is a variety of approaches for imitation learning and many of them have means for handling uncertainty in the underlying models~\cite{ROB-053}. For instance, Ziebart \emph{et al.}~\cite{Ziebart2008:AAAI} use inverse reinforcement learning to learn a cost function from demonstrations, and is thus able to cope with various sources of uncertainty, including in input features. 

End-to-end behavioral cloning is a popular approach for learning to drive from demonstrations~\cite{Pomerleau1988,LeCun2005driving,Bojarski2016nvidiadriving,Codevilla2018,Wang18,cilrs2019iccv}. However, these methods suffer from generalization issues~\cite{cilrs2019iccv} that require potentially unsafe on-policy corrections~\cite{Ross2011dagger}, although maybe in a limited amount~\cite{ZhangCho2017}.
Beyond raw sensors, one can use high-level perceptual representations for learning from demonstrations~\cite{wulfmeier2017large,muller2018driving, wang2019mpv, bansalchauffeurnet2019, zhou2019vision}. In particular, with perfect mapping, localization, and perception one can learn complex urban driving via imitation~\cite{chen2019learning}. 
While visual abstractions lend themselves to sample-efficient  imitation learning, their use does not capture uncertainty and noise can lead to degraded performance. In this work we propose an approach to overcome this fundamental limitation.

Representations for uncertainty in Deep Learning are a topic of surging interest~\cite{kendall17uncertainty}, with both variational~\cite{gal2016dropout,kingma2013auto} and adversarial approximations~\cite{goodfellow2014generative} employed to represent uncertainty. More recent work~\cite{gast2018lightweight} has shown that for certain types of networks the model uncertainty can even be computed in closed form. This opens the opportunity to get uncertainty estimates in real-time.
In perception systems, several recent works incorporate uncertainty reasoning and deep representations~\cite{feng2018towards,meyer2019lasernet}.
While some approaches in imitation learning attempt to be aware of the limitations of the system~\cite{Richter2017,amini2018variational,huang2019uncertainty}, uncertainty reasoning to make the system more robust to uncertain \textit{inputs} is still relatively limited, and is the focus of our work.
Czarnecki and Podolak~\cite{czarnecki2013uncertainty} present a training scheme for handling known input data uncertainties to improve generalizability, but they do not show how the approach can be used with CNNs or more complex tasks like path planning.
\\
Compared to the approaches described above we in this paper focus specifically on how to deal with false positives due to perception errors in the context of automated driving through behavioral cloning, where the expert has access to the ground truth while the robot might suffer from false positive perceptions. 

\section{Imitation under Perceptual Uncertainty}

In this section, we present our behavioral cloning algorithm that can effectively leverage mediated perception even in the presence of false positives.

\subsection{Imitation Learning with Mediated Perception}

Our goal is to learn the parameters $\params$ of a policy $\pi$ that can predict robot actions $\act$ from observations $\obs$, i.e., $\pi(\obs; \params) = \act$.
In our case, robot actions are future way-points $\act=\{w_1, \ldots w_K\}$ passed to a downstream controller. In contrast to end-to-end sensorimotor approaches, we assume observations are not raw sensor signals but instead the outputs of a perception system (e.g., object tracks, localization and mapping information).

We estimate the policy parameters $\theta$ by behavioral cloning, i.e., by supervised learning from a set of optimal demonstrations $\{\traj_1, \ldots, \traj_n\}$ generated by an expert policy $\pi_e$. Although the expert (a.k.a.\ oracle) has access to the true world state $\state$ (e.g., ground truth position of other agents), this high-level information cannot be measured directly by the robot's sensors.
Therefore, we assume each demonstration $\tau_k$ consists of observation-action pairs $\tau_k = (\obs_k, \act_k)$ where observations are \textit{recorded predictions of the robot's perception system}. We, however, assume that the actions $\act$ can be accurately measured (e.g., using GNSS systems or recording CAN bus signals for cars).
This setup is different than standard behavioral cloning~\cite{Pomerleau1988, Schaal2003imitation, Ross2011dagger, levine2017learning, Codevilla2018, cilrs2019iccv}, as we are trying to approximate an expert while operating over a different input space:
\begin{equation}
    \pi(\obs; \params) \sim \act = \pi_e(\state)
\end{equation}
A key challenge lies in potential inconsistencies between observations $\obs$ and the true state $\state$, for instance in the presence of false positives in $\obs$. This is incompatible with supervised learning, as the same observations may yield potentially different target actions (e.g., stopping or passing through an obstacle).

\subsection{Input Uncertainty Representation via Soft BEV}
\label{sec:input_uncertainty}

To overcome inconsistencies between observations and actions, we leverage uncertainty estimates provided by modern perception systems. We model observations $\obs = (\hat{\state}, \conf)$ as pairs of estimated perceptual states $\hat{\state}$ and black-box confidence values $\conf$ in $[0; 1]$ for each state variable.
We assume these confidence estimates negatively correlate with error rates of the respective perception sub-systems, although they might be inaccurate. This is a reasonable assumption in practice, as there are multiple ways to achieve this, for instance via Bayesian Deep Learning~\cite{kendall17uncertainty}, ensemble methods~\cite{lakshminarayanan2017simple}, or parametric approximation~\cite{meyer2019lasernet,huang2019uncertainty}.

We do not make explicit assumptions about the distribution of $\hat{\state}$ w.r.t.\ the true state $\state$. Instead, we assume the perception system is tuned for \textit{high recall}, i.e., that all critical state variables are (noisily) captured in the estimated state. This comes at the potential expense of false positives, but corresponds to the practical setup where safety requirements are generally designed to avoid partial observability issues, as false negatives are hard to recover from.
More specifically, we tackle the inherent underlying noise in the perception system and not the issue of how to handle completely out-of-distribution cases (anomalies) in the perception system, where the confidence estimates are no longer positively linked to the true error rates.

We represent the observations $\obs$ in a birds-eye-view grid, i.e., an $N{\times}M{\times}D$-dimensional tensor $\bev$ where each dimension $k$ represents a category of estimated state (e.g., an object or feature type) together with the respective estimated confidences. Each slice $\bev_k$ is a matrix $\bev_k \in [0;1]^{N \times M}$, where each element corresponds to the presence of an estimated object or feature of type $k$ at that location, weighted by its estimated confidence.
%
%
We refer to the resulting input representation $\bev$ as \emph{Soft BEV}.

\begin{figure}
\centering
\vspace*{4mm}
\includegraphics[width=0.48\textwidth]{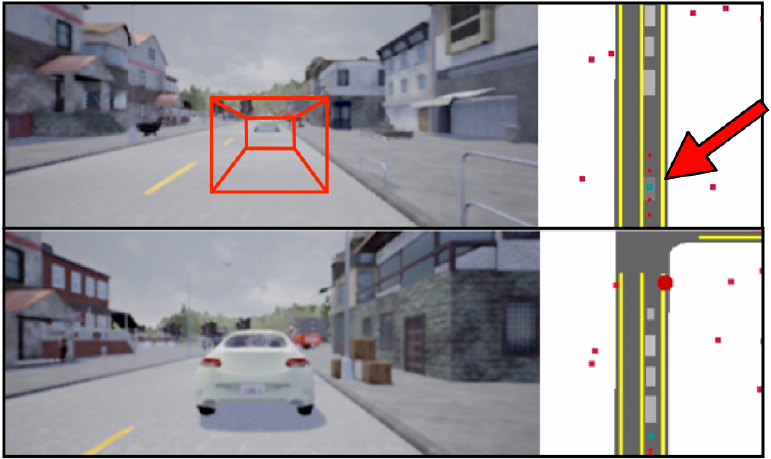}
\caption{Qualitative illustration of an agent using the \emph{Soft BEV} grid, successfully ignoring a false positive (top). Note that a few instances later it is able to stop for a true positive (bottom)}
\label{fig:qualitative}
\end{figure}
\subsection{Behavioral Cloning with Soft BEV}
\label{sec:system_setup}

We model a driving agent via a deep convolutional policy network taking as input the aforementioned Soft BEV representation $\bev$.
The CNN outputs way-points along the future trajectory, which are then used by a PID controller to compute the control signals for the steering and throttle of the vehicle.
We use the same network architecture as proposed by Codevilla \emph{et al.}~\cite{cilrs2019iccv} and Chen \emph{et al.}~\cite{chen2019learning}.
It consists of a ResNet-18~\cite{he2016deep} base network acting as an encoder, followed by three deconvolutional layers which also have as an input the current speed signal. 
For each of the potential high-level commands (``go left'', ``go right'', ``go straight'', ``follow the road''), the network predicts multiple output heat-maps which are then converted into way-points by spatial soft-argmax layers. 
Based on the high-level command, the respective head of the network is used to predict the way-points.

To operate correctly under uncertainty, the goal is to learn a policy that fulfills the standard behavioral cloning target, while additionally remaining invariant to the noise in the input features.
%
%
As we directly encode uncertainty in the input representation, the optimal actions (as done by the expert) can still be optimal under perceptual noise, as long as the behavioral patterns are not dominated by a wrong bias in uncertainty estimates. This is a reasonable assumption in practice, as consistent patterns of errors can be characterized on a validation set and addressed specifically. 
Therefore, the problem can still be treated as a supervised policy learning~\cite{Ross2011dagger,Bojarski2016nvidiadriving}, solving the following optimization problem:
\begin{equation}
    \params^* = \argmin_{\params}{\sum_{i}{l(\pi(\bev^i; \theta), \pi_{e}(\state_i))}}
\end{equation}
where $l$ is a loss function, in our case the $L_1$-distance.

\section{Experiments}
\begin{figure}
\centering
\vspace*{4mm}
\includegraphics[width=0.48\textwidth]{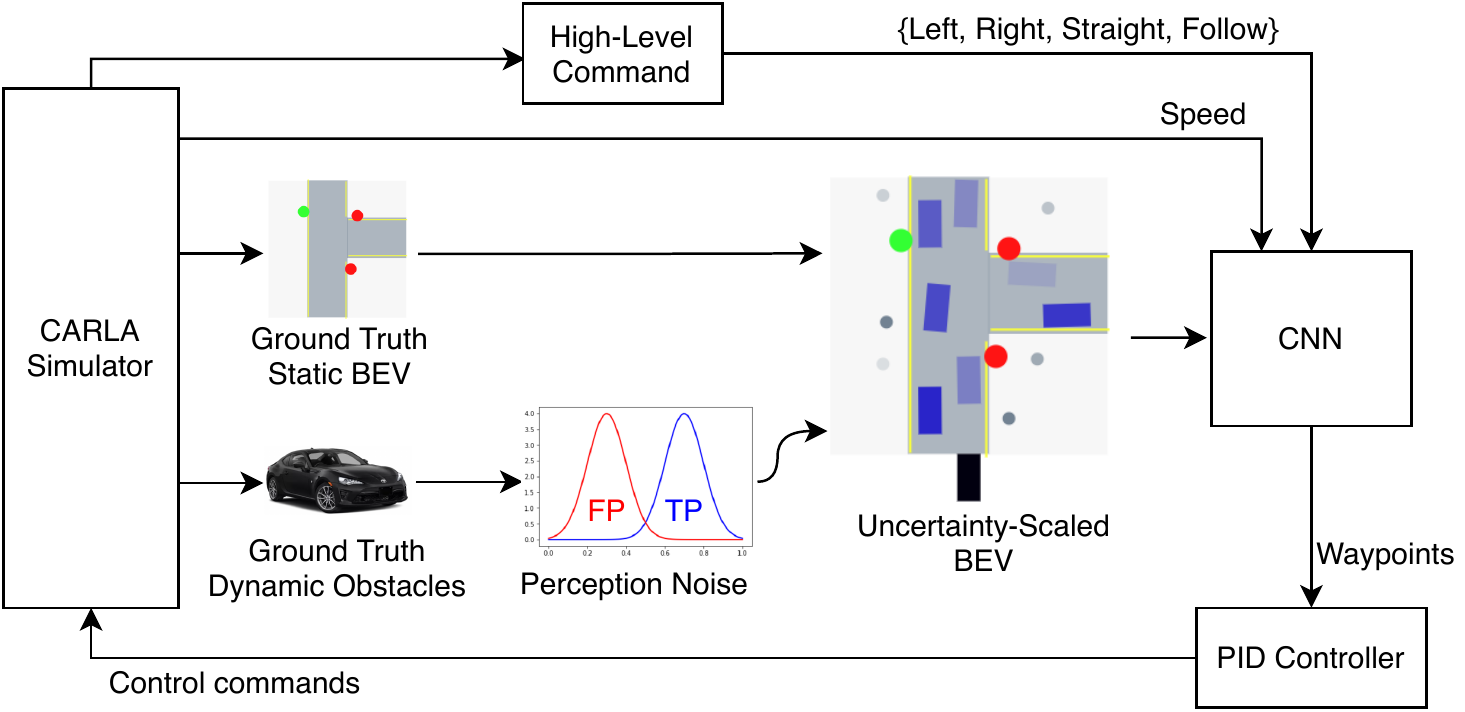}
\caption{Experimental setup: The CARLA simulator provides ground truth features. Perception noise is applied to the dynamic features, which are then fused into an uncertainty-scaled birds-eye view representation, the \emph{Soft BEV}. Together with high-level commands and speed information it is fed to a CNN that predicts way-points.}
\label{fig:system_setup}
\end{figure}
In this section we provide experimental evidence that the incorporation of uncertainty into the input representation of an imitation learning based planner can improve its performance. 
For this we use the autonomous driving simulator CARLA (version 0.9.6). 
The simulator provides ground truth road layout, lane lines, traffic light state and dynamic objects in the surroundings of the ego vehicle.
A front view of the agent in the simulation environment is depicted in Fig.~\ref{fig:qualitative}.

\subsection{Experimental Setup}
We describe the experimental setup, which includes the dataset, benchmarks and the training procedure for the CNN.
\subsubsection{Dataset}
We use a rule-based autopilot to collect a dataset of 175k training frames and 34k validation frames in CARLA's ``Town1''. 
The scenarios are created by having the agent navigate along pre-defined way-points through the simulated city.
The collected data contains the following information at each frame: global position, rotation, velocity, high-level navigation command ({left, right, straight, follow}), birds-eye view road layout, lane markings and traffic lights, and a list of dynamic obstacles. 
The main attributes given for the dynamic obstacles are the position in local coordinate frame, relative rotation and bounding box extent.  

\subsubsection{Benchmark}
The performance of our agents is evaluated on the NoCrash Benchmark proposed by Codevilla \emph{et al.}~\cite{cilrs2019iccv}. 
More specifically, we perform evaluations in the ``Town2 - Regular Traffic'' setting, where a moderate number of cars and pedestrians are present. 
We exclude the ``Town2 - Empty Traffic'' and ``Town2 - Dense Traffic'' settings, since the former does not include any dynamic objects, and the ladder does in our experience often not provide meaningful results due to unwanted side effects like traffic jams or unavoidable collisions with pedestrians. 
The benchmark performs multiple runs, where each run is outlined by a fixed starting position, and a target goal position. 
The performance of an agent is evaluated with regards to multiple criteria.
Firstly, if the agent crashes into another dynamic or static object, the run is considered as failed. 
Secondly, if the agent reaches the target location the time to completion is captured. 
Thirdly, if the agent is not able to finish the run within a certain time limit, the run is considered as failed as well. 
This includes the cases in which the agent takes a wrong turn at one point, or when it remains stuck somewhere.
Lastly, traffic light violations are counted, but in the scope of our analysis not relevant for the evaluation process.
It is to be noted that we do not distinguish between the different weather modes in the benchmark, since our evaluations never use visual (camera) input.
\subsubsection{Training Setup}
We train two agents, both using the exact same network architecture.
The only difference between them lies in the way the input data is represented.
The first agent (BEV Agent) is trained on ground truth noise-free data, meaning it will have no notion of uncertainty.
The second agent (Soft BEV Agent) is trained on noisy data and its input is the \emph{Soft BEV grid}.

At training time we perform the following procedure to simulate the output of a perception stack, i.e., to simulate noise.
We use two truncated Normal distributions (values between 0 and 1) as shown in Fig.~\ref{fig:system_setup} (``Perception Noise'') to simulate the confidence score generated by a generic perception stack, and approximate the perception system's confidence estimate on its failure modes:
\begin{equation}
    c_j \thicksim \mathcal{N}_{trunc}(\mu_j, \sigma_j), j \in [FP, TP]
\end{equation}
The distributions are designed to simulate the uncertainty of the output of a statistically well calibrated perception system, where this uncertainty can be interpreted as a probability of existence for an object.
The first distribution (FP) is used to sample confidence scores for false positives and is biased towards lower scores, while the second distribution for true positives (TP) is biased towards higher scores.
After sampling true positive confidence values for the actual objects, we sample ghost-objects in the vicinity of the ego agent with a probability of $10\%$ and sample a confidence value for this false positive object as well. 
Then, we incorporate these objects into the tensor following our approach introduced in Section~\ref{sec:input_uncertainty}. Following~\cite{chen2019learning,Codevilla2018} we also augment the input BEV image by rotation, shifting and cropping. To simulate trajectory noise, the way-point labels are shifted and rotated.


\subsubsection{Testing Setup}
To model temporal consistency of the false positives we employ a survival model. 
In other words a false positive is created with probability $p_{FP}$, and it will be present in the consecutive frames with a decaying probability. In our experiments, we set the decay rate to $0.8$, thus allowing false positives to typically survive 2-3 frames.
For each frame, we sample for every surviving false positive a score from the respective truncated Normal distribution.
Next, we create the Soft BEV grid as described previously.
Figure~\ref{fig:system_setup} depicts the setup used to perform the experiments. 

\subsubsection{3D Object Detector} 
To test our agents in a more realistic scenario, we implemented a LiDAR based 3D object detector based on the state-of-the-art PointPillars (PP)~\cite{lang2019pointpillars} approach. Briefly, the detector first performs an initial voxelization step (we use a grid of resolution  $0.25m \times 0.25m$). Each voxel is then processed by a multi-layer perceptron~\cite{qi2017pointnet} to create an intermediate Birds-Eye-View representation. Finally, the 3D detections are computed by a Single-Shot Detector (SSD)~\cite{liu2016ssd} head. We used CARLA to simulate LiDAR sensors with 32 and 64-beams respectively, and we trained the detector on 100k 360 degrees LiDAR sweeps and validated on 10k LiDAR sweeps.

\begin{figure}
\centering
\vspace*{2mm}
\includegraphics[width=0.48\textwidth]{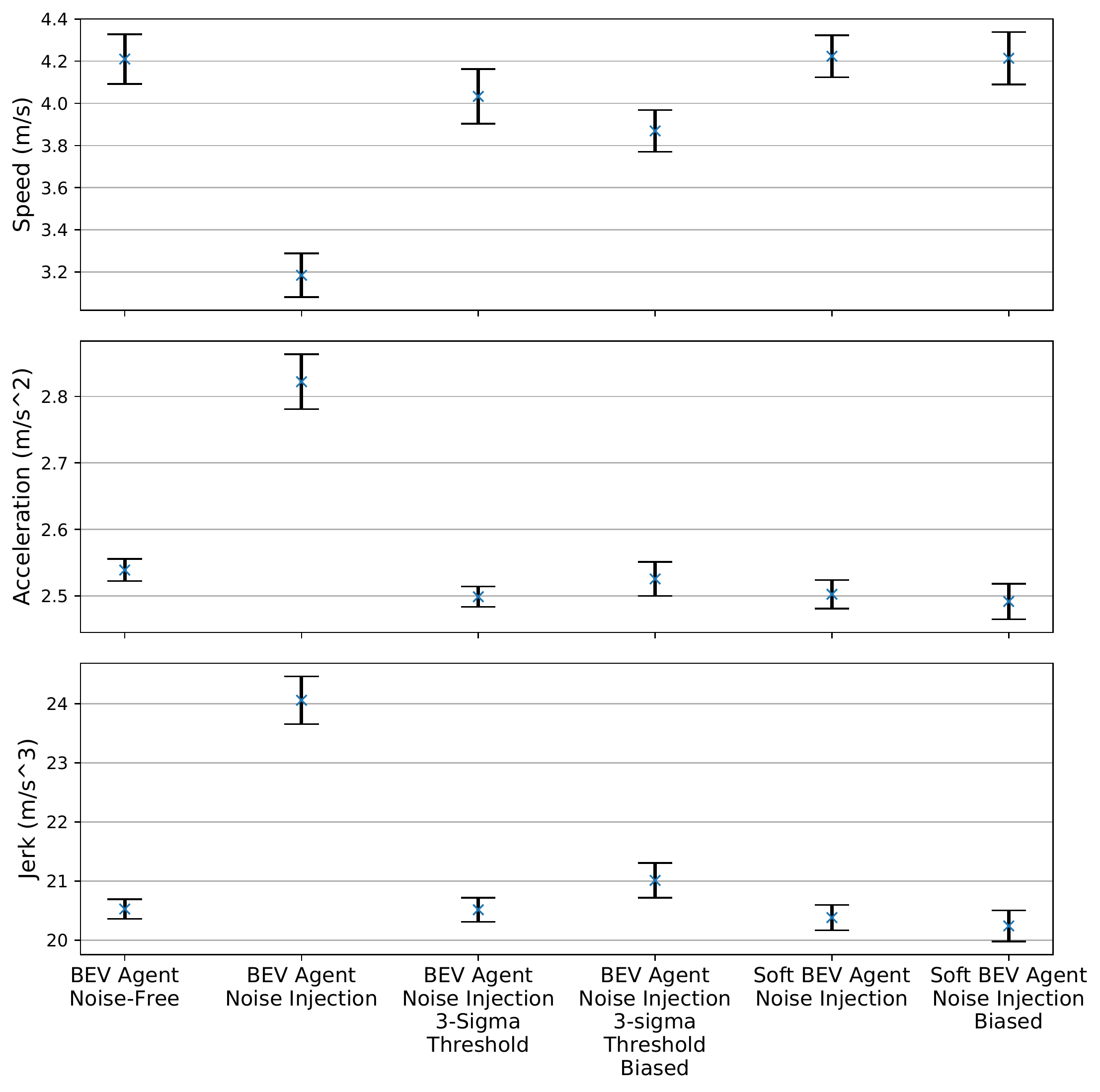}
\caption{Average speed and average absolute acceleration and jerk of the BEV Agent and the Soft BEV agent in multiple noise settings for the NoCrash benchmark. We plot the mean difference and corresponding 95\% confidence interval computed using a t-test, where each configuration is compared against the ``BEV Agent Noise-Free''.}
\label{fig:results_plot}
\end{figure}

\subsection{Results}
We perform multiple experiments to test the efficacy of the introduced \emph{Soft BEV grid}.
More specifically we want to evaluate if it is possible to reduce conservative behavior in uncertain situations, while simultaneously acting safely.
\subsubsection{Comparison BEV vs Soft BEV}

\begin{table*}[ht]
\vspace*{2mm}

\centering
\arrayrulecolor[rgb]{0.8,0.8,0.8}
\begin{tabular}{|c|l|c|c|c|c|c|c|c|c|} 
\arrayrulecolor{black}\hline
\multicolumn{1}{|c!{\color{black}\vrule}}{Config.} & \multicolumn{1}{l!{\color{black}\vrule}}{Model} & \multicolumn{1}{c!{\color{black}\vrule}}{Noise} & \multicolumn{1}{c!{\color{black}\vrule}}{3-Sigma threshold} & \multicolumn{1}{c!{\color{black}\vrule}}{p-existence} & \multicolumn{1}{c!{\color{black}\vrule}}{Bias [Mu, Sigma]} & \multicolumn{1}{c!{\color{black}\vrule}}{Speed} & \multicolumn{1}{c!{\color{black}\vrule}}{Acceleration} & \multicolumn{1}{c!{\color{black}\vrule}}{Jerk} & \multicolumn{1}{c!{\color{black}\vrule}}{Collision}  \\ 
\hline
1                                                  & BEV Agent                                       & x                                               & Not applicable                                              & Not applicable                                        & Not applicable                                             & 4.2                                             & 2.5                                                    & 20.5                                           & 0\%                                                  \\ 
\arrayrulecolor[rgb]{0.8,0.8,0.8}\hline
2                                                  & BEV Agent                                       & \checkmark                                               & x                                                           & Low, 0.1                                              & x                                                          & 3.2                                             & 2.8                                                    & 24.1                                           & 0\%                                                  \\ 
\hline
3                                                  & BEV Agent                                       & \checkmark                                               & \checkmark                                                           & Low, 0.1                                              & x                                                          & 4.0                                             & 2.5                                                    & 20.6                                           & 5\%                                                  \\ 
\hline
4                                                  & BEV Agent                                       & \checkmark                                               & \checkmark                                                           & Low, 0.1                                              & High, [0.2, 0.1]                                           & 3.9                                             & 2.5                                                    & 21.0                                           & 5\%                                                  \\ 
\hline
5                                                  & Soft BEV Agent                                  & \checkmark                                               & x                                                           & Low, 0.1                                              & x                                                          & 4.2                                             & 2.5                                                    & 20.4                                           & 5\%                                                  \\ 
\hline
6                                                  & Soft BEV Agent                                  & \checkmark                                               & x                                                           & Low, 0.1                                              & High, [0.2, 0.1]                                           & 4.2                                             & 2.5                                                    & 20.3                                           & 5\%                                                  \\
\hline
\end{tabular}
\arrayrulecolor{black}

\caption{Comparison of different agents in different noise configurations on the NoCrash Benchmark. We report the performance metrics average speed, average absolute acceleration, average absolute jerk and total collision rate.}
\label{table:results_summary}
\end{table*}
We summarize the findings of our first set of experiments in Table~\ref{table:results_summary} and
Figure~\ref{fig:results_plot} depicts the results visually. 
%
%
They show the mean difference and 95\% confidence intervals of different experiments when evaluating the metrics: speed, acceleration and jerk using a t-test.
To reduce the influence of traffic lights, we only consider velocities above $1m/s$ when computing the average velocity.
The reason for this is that traffic lights act as a binary gate and essentially either increase variance significantly, or eliminate the influence of speed difference along the routes.
Similarly, we only consider accelerations with an absolute value above zero.
We compare the two agents in six different configurations to analyze the influence of the appearance of false positives.
The model named BEV Agent refers to an agent that has been trained on noise-free data, and has no notion of uncertainty.
The Soft BEV Agent has been trained using noise injection and its input is represented using the Soft BEV. 
We compare the different configurations in average speed, average absolute acceleration, average absolute jerk and total collision rate over all the benchmark runs.
These metrics can be used to interpret how conservative the trained agent is, while remaining safe.
In addition to the two agents, we look at the following configuration parameters:
\begin{itemize}
\item 3-Sigma threshold: A 3-sigma threshold is applied on the detection uncertainties and all values below the threshold are discarded.
\item p-existence: Probability of occurrence of a false positive object within the relevant field of view for the agent. See Section~\ref{sec:fov} for the respective analysis regarding influence of field of view on agent performance.
\item Biased: A bias is applied to the mean $\mu_j$ and standard deviation $\sigma_j$ of the underlying false positives and true positives normal distribution. The bias itself is sampled from a uniform distribution:
\begin{equation}
    \mu^{bias}_j = \mu_j + b_{\mu_j}
\end{equation}
\begin{equation}
    b_{\mu_j} \thicksim \emph{Uniform}(-b_{\mu}, b_{\mu})
\end{equation}
and
\begin{equation}
    \sigma^{bias}_j = \sigma_j + b_{\sigma_j}
\end{equation}
\begin{equation}
    b_{\sigma_j} \thicksim \emph{Uniform}(-b_{\sigma}, b_{\sigma})
\end{equation}
where $\mu^{bias}_j$ and $\sigma^{bias}_j$ represent the new mean and standard deviation.
\end{itemize}
Configuration 1 is an upper bound on the performance, since it is evaluated on the ground truth BEV input representation and therefore has perfect knowledge of the environment.
The main result of this evaluation can be seen in Configuration 5, where from the noisy data the Soft BEV input representation is able to recover close to full performance. 
Additionally, as seen in Configuration 6, this is possible even in the case of a biased underlying false positives distribution.
This means that the agent is robust even when misestimations of the true noise distribution occur. 
Configurations 3 and 4 depict the case when a 3-sigma threshold is applied to the noisy observations.
In this case in which the underlying distribution parameters are perfectly known (Configuration 3), the performance can be recovered.
In the more realistic case, where the distribution parameters are biased, the average speed drops and the absolute average acceleration and jerk increase, demonstrating that a naive thresholding is not enough.
The second configuration shows that the influence of the injected noise is very significant, if the uncertainty value is completely ignored.
Only the noise free configurations 1 and 2 are able to complete the benchmarks collision free, although the statistical significance of this metric is low. 
The data for these experiments has been evaluated using an unequal variance, unequal sample sizes t-test where each configuration is compared against configuration 1. 
\subsubsection{Influence of noise on Soft BEV}
\begin{table*}
\vspace*{2mm}
\centering
\arrayrulecolor[rgb]{0.8,0.8,0.8}
\begin{tabular}{|l|c|c|c|c|c|} 
\arrayrulecolor{black}\hline
\multicolumn{1}{|l!{\color{black}\vrule}}{Model} & \multicolumn{1}{c!{\color{black}\vrule}}{Noise Injection} & \multicolumn{1}{c!{\color{black}\vrule}}{3-Sigma Threshold} & \multicolumn{1}{c!{\color{black}\vrule}}{p-existence} & \multicolumn{1}{c!{\color{black}\vrule}}{Bias [Mu, Sigma]} & \multicolumn{1}{c!{\color{black}\vrule}}{Benchmark Success Rate}  \\ 
\hline
Soft BEV Agent                                   & \checkmark                                                         & x                                                           & Low, 0.05                                             & x                                                          & 97.5\%                                                            \\ 
\arrayrulecolor[rgb]{0.8,0.8,0.8}\hline
Soft BEV Agent                                   & \checkmark                                                         & x                                                           & Medium, 0.1                                           & x                                                          & 92.5\%                                                            \\ 
\hline
Soft BEV Agent                                   & \checkmark                                                         & x                                                           & High, 0.5                                             & x                                                          & 95.0\%                                                            \\ 
\hline
Soft BEV Agent                                   & \checkmark                                                         & x                                                           & Low, 0.05                                             & Low, [0.01, 0.005]                                         & 92.5\%                                                            \\ 
\hline
Soft BEV Agent                                   & \checkmark                                                         & x                                                           & Low, 0.05                                             & Medium, [0.05, 0.025]                                      & 90.0\%                                                            \\ 
\hline
Soft BEV Agent                                   & \checkmark                                                         & x                                                           & Low, 0.05                                             & High, [0.2, 0.1]                                           & 92.5\%                                                            \\
\hline
\end{tabular}
\arrayrulecolor{black}

\caption{Influence of frequency of false positives and bias on the confidence score.}
\label{table:results_bias}
\end{table*}
In the second set of experiments we investigate the influence of the frequency of false positives and of bias on the distribution parameters.
These experiments are motivated by the fact that in a real-world application, the underlying distribution parameters are almost always unknown, and even potentially change over time.
This experiment induces a gap between training and testing noise characteristics, which is very typical in a realistic scenario. 
The results can be seen in Table~\ref{table:results_bias}. 
The \emph{Soft BEV} agent’s performance is weakly affected by an increase in the number of false positives, or a change in the underlying distribution. 
Even under more radical changes the performance does not deteriorate much further.
We therefore show that our \emph{Soft BEV} model is able to maintain the same level of safety despite not knowing the true underlying noise distribution and even when this distribution diverges from the training distribution, which is the more realistic scenario. Additionally we show that we do not suffer any deterioration in the quality of driving despite the added noise.
\begin{table} [h!]

\centering
\arrayrulecolor[rgb]{0.8,0.8,0.8}
\begin{tabular}{|l|c|c|c|c|c|c|} 
\hline
\multirow{2}{*}{3D-Detector} & \multicolumn{3}{c|}{$AP_{Car}$} & \multicolumn{3}{c|}{$AP_{Ped}$} \\ 
& 0.5m & 1.0m & 2.0m & 0.5m & 1.0m & 2.0m \\
\hline
\multicolumn{7}{|c|}{Objects closer than 25m} \\ 
\hline
PP - 32 beams & 90.7 & 92.8 & 95.8 & 58.8 & 60.6 & 61.8 \\
\hline
PP - 64 beams & 84.0 & 90.5 & 93.2 & 60.1 & 68.1 & 69.4 \\
\hline

 \multicolumn{7}{|c|}{Objects closer than 50m} \\ 
\hline
PP  - 32 beams & 65.5 & 69.6 & 74.5 & 27.4 & 28.1 & 29.0 \\
\hline
PP  - 64 beams & 62.2 & 72.6 & 75.8 & 38.9 & 44.0 & 45.22 \\
\hline

\end{tabular}
\arrayrulecolor{black}

\caption{PointPillars (PP)~\cite{lang2019pointpillars} 3D Detector Results on CARLA LiDAR data. We report the nuScenes~\cite{caesar2019nuscenes} Average Precision (AP) metric on Cars and Pedestrians.}
\label{table:3ddet_nuscenes_metrics}
\end{table}
%
%
\subsubsection{Soft BEV with 3D Detector}
In the last set of experiments we want to see how well the agents work when the dynamic observations are perceived by a LiDAR-based perception system.
We first validate the performance of our PointPillars LiDAR detector on Cars and Pedestrians by reporting the nuScenes~\cite{caesar2019nuscenes} Average Precision (AP) metrics on the validation set (Table~\ref{table:3ddet_nuscenes_metrics}). Our detectors perform quite well on cars, and we note a significant accuracy drop for pedestrians for detections up to $50m$. This can be attributed to the quality and density of the CARLA LiDAR simulation (4k points for the 32-beam LiDAR and 9k points for the 64 beam LiDAR). We note that the numbers we report are consistent with those of state-of-the-art 3D object detection methods on the nuScenes 3D object detection benchmark~\cite{caesar2019nuscenes} from which we conclude that our implementation allows us to test our agents with a realistic perception system. 

The goal of this experiment is to test the flexibility of our \emph{Soft BEV} representation, since the distribution of the confidence scores is unknown to the agent.
\begin{table}

\centering
\arrayrulecolor[rgb]{0.8,0.8,0.8}
\begin{tabular}{|l|c|c|} 
\hline
                                                       & \multicolumn{2}{c|}{Collision Rates}                                                                                                                                                                                                 \\ 
\hline
\multicolumn{1}{|c!{\color{black}\vrule}}{3D-Detector} & \multicolumn{1}{l!{\color{black}\vrule}}{\begin{tabular}[c]{@{}l@{}}BEV Agent with \\0.5-Threshold\end{tabular}} & \multicolumn{1}{c!{\color{black}\vrule}}{\begin{tabular}[c]{@{}c@{}}Soft BEV Agent, \\no Threshold\end{tabular}}  \\ 
\arrayrulecolor{black}\hline
PP - 32 beams                                     & 10\%                                                                                                             & 5\%                                                                                                               \\ 
\arrayrulecolor[rgb]{0.8,0.8,0.8}\hline
PP - 64 beams                                     & 5\%                                                                                                              & 5\%                                                                                                               \\
\hline
\end{tabular}
\arrayrulecolor{black}

\caption{Collision rates when using the PointPillars (PP)~\cite{lang2019pointpillars} 3D detector.}
\label{table:result_3ddetector}
\vspace*{-4mm}
\end{table}
We compare against a BEV agent that thresholds the observations at 0.5 and show the results of this experiment in Table~\ref{table:result_3ddetector}.
We find that our method outperforms an engineered approach and achieves a lower collision rate in the case of the 32-beam LiDAR. 
Our method as well as the engineered approach perform equally well when using the 64-beam LiDAR detector, which can be attributed to the fact that the detection rate of this model is quite high on CARLA data, especially for objects closer than $25m$. 
As we show in the next section, for the purposes of the benchmarks considered in this work, even this limited range is enough to recover full performance. 
%


\subsubsection{Field of View Experiments}
\begin{figure}
\centering
\includegraphics[width=0.35\textwidth]{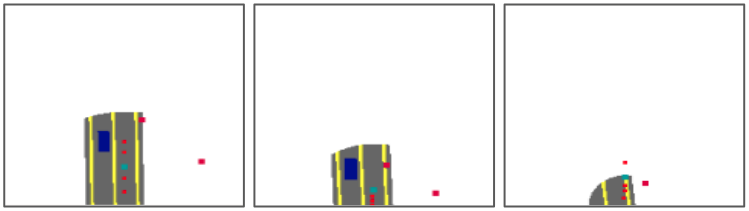}
\caption{Depiction of the reduced field of view.}
\label{fig:fov}
\end{figure}

We analyzed the influence of the agent's field of view (FoV) on its performance.
For this we reduce the FoV to a circular section centered around the agent, as illustrated in Figure~\ref{fig:fov}.
Interestingly, the agent is able to recover full performance ($97.5\%$ success rate) when the FoV is reduced to a circle spanning less than half of the total FoV, which corresponds to a radius of $18m$. 
As the FoV is reduced further the performance drops to $57.5\%$ successful runs (radius $12m$), and finally it reaches $0\%$ when reducing it to a half-circular region with radius of $6m$.

\label{sec:fov}

\section{Conclusion}
In this work we propose an effective way to deal with false positives for behavioral cloning of autonomous driving policies with mediated perception. 
To achieve this, we introduced the probabilistic birds-eye-view (\emph{Soft BEV}) grid and showed how this simple representation of perceptual uncertainty can be used to better deal with false positives without the need for thresholds. 
In extensive experiments carried out with the CARLA simulator, we showed that we can recover noise-free, safer driving policies from expert demonstrations, even when uncertainty estimates might be themselves noisy.
%


\addtolength{\textheight}{-0cm}   



\section*{ACKNOWLEDGMENT}
We express our deepest gratitude to Prof. Roland Siegwart from the Autonomous Systems Lab for initiating and facilitating this collaboration between ETH Zurich and Toyota Research Institute (TRI). This research was supported by TRI and the article solely reflects the opinions and conclusions of its authors and not TRI or any other Toyota entity.

\bibliographystyle{IEEEtran}
\bibliography{end-to-end-driving}

\end{document}